%% file: neurips_2026.tex
\documentclass{article}

\usepackage[preprint]{neurips_2026}

\usepackage[utf8]{inputenc} % allow utf-8 input
\usepackage[T1]{fontenc}    % use 8-bit T1 fonts
\usepackage{hyperref}       % hyperlinks
\usepackage{url}            % simple URL typesetting
\usepackage{booktabs}       % professional-quality tables
\usepackage{amsfonts}       % blackboard math symbols
\usepackage{nicefrac}       % compact symbols for 1/2, etc.
\usepackage{microtype}      % microtypography
\usepackage{xcolor}         % colors
\usepackage{multirow}
\usepackage[ruled,vlined]{algorithm2e}
\usepackage{graphicx}
\usepackage{amsmath}
\usepackage{amssymb}
\usepackage{enumitem}

\title{MemAudit: Post-hoc Auditing of Poisoned Agent Memory via Causal Attribution and Structural Anomaly Detection}

\author{
 \textbf{Zhewen Tan\textsuperscript{1,2,3,4}},
 \textbf{Yilun Yao\textsuperscript{2,3}},
 \textbf{Huiyan Jin\textsuperscript{3}},
 \textbf{Wenhan Yu\textsuperscript{2,3}},
 \textbf{Guoan Wang\textsuperscript{3}},
 \\
 \textbf{Mengyuan Fan\textsuperscript{3}},
 \textbf{liang lu\textsuperscript{1,4}},
 \textbf{Feng Liu\textsuperscript{1,4}},
 \textbf{Xiangzheng Zhang\textsuperscript{2}},
 \\
 \textbf{Duohe Ma\textsuperscript{1,4}\thanks{Corresponding author.}},
 \textbf{Tong Yang\textsuperscript{3}\footnotemark[1]},
 \textbf{Lin Sun\textsuperscript{2}\footnotemark[1]}
\\
\textsuperscript{1}Institute of Information Engineering, Chinese Academy of Sciences \quad
\textsuperscript{2}Qiyuan Tech \\
\textsuperscript{3}Laboratory of Multimedia Information Processing, School of Computer Science, Peking University
\\
\textsuperscript{4}School of Cyber Security, University of Chinese Academy of Sciences
\\
 \small{
   \textbf{Correspondence:} Duohe Ma: \href{mailto:maduohe@iie.ac.cn}{maduohe@iie.ac.cn}, Tong Yang: \href{mailto:yangtong@pku.edu.cn}{yangtong@pku.edu.cn}, Lin Sun: \href{mailto:sunlin1@360.cn}{sunlin1@360.cn}
 }
}

\begin{document}

\maketitle
\input{0_abstract}

\input{1_Introduction}
\input{2_Related_Work}
\input{3_Preliminary}
\input{4_Methodology}
\input{5_Experimental_Setup}
\input{6_Results}
\input{7_Conclusion}
\input{8_Limitations}

\bibliographystyle{plainnat}
\bibliography{references}

\appendix

\input{9_Appendix}

\newpage
\input{Checklist}
\end{document}

%% file: 0_Abstract.tex
\begin{abstract}
    Large language model agents increasingly rely on persistent memory to store past interactions, retrieve relevant demonstrations, and improve long-horizon task execution. However, this memory mechanism also creates a practical security vulnerability: an adversarial user may inject malicious records into the agent's memory through ordinary interaction, and these records can later be retrieved to steer the agent's reasoning and actions. Existing defenses primarily focus on online intervention, such as prompt filtering or output blocking, but they do not address the post-hoc question of which stored memories are responsible after harmful behavior has already been observed. We propose \textbf{MemAudit}, a post-hoc causal memory auditing framework for memory-augmented LLM agents. The framework combines two complementary signals: (1) a counterfactual memory influence score that measures each memory's causal contribution to harmful outputs, and (2) a memory consistency graph that identifies structurally anomalous memories within the broader memory store. We evaluate MemAudit against MINJA, a query-only memory injection attack in which malicious records are generated and stored through normal agent interactions rather than direct memory-bank modification. Across both QA and reasoning-agent settings, MemAudit substantially reduces attack success rates under realistic post-hoc auditing scenarios. The results show that QA attack success is reduced from $70\%$ to $0\%$, while RAP attack success drops from $83.3\%$ to $0\%$.
\end{abstract}

%% file: 1_Introduction.tex
\section{Introduction}
\label{sec:intro}

Large language model (LLM) agents are rapidly evolving from passive conversational assistants into autonomous systems capable of interacting with external environments and executing complex long-horizon tasks~\cite{yao2022react}. Recent agent systems such as OpenHands demonstrate that modern agents can already perform realistic software engineering, web interaction, command execution, and multi-step planning tasks, enabling their increasing integration into daily workflows such as software development, personal assistance, information management, and online automation~\cite{wang2024openhands}. As these systems become more capable, users are also beginning to delegate increasingly sensitive permissions and decision-making authority to agents. 

This trend substantially raises the security stakes of autonomous agents. Once an agent performs an unsafe or manipulated action, the resulting damage may extend far beyond a single incorrect text generation. Harmful actions can cause severe real-world consequences by propagating harmful decisions across external systems and long-running user workflows~\cite{ferrag2025prompt}. As a result, ensuring the reliability and safety of autonomous agents is becoming a critical challenge for real-world deployment.

A major factor behind modern agent capability is the use of persistent memory~\cite{lewis2020retrieval}. Many recent agents continuously store observations, reasoning trajectories, user preferences, and historical interactions to support adaptive decision making across sessions~\cite{shinn2023reflexion}. External memory enables agents to accumulate experience over time and support long-horizon adaptation instead of operating as stateless generators~\cite{wang2023voyager}. This design significantly improves long-horizon reasoning, personalization, and behavioral consistency~\cite{park2023generative}. However, the mechanism also introduces a dangerous new attack surface: once malicious information is written into memory, it can remain active across future interactions and repeatedly influence the agent's behavior.

Recent work has shown that such memory attacks are practical in deployed agent systems. MINJA demonstrates that adversaries can inject malicious reasoning trajectories into agent memory through natural interaction, causing delayed and persistent behavioral manipulation across future scenario~\cite{dong2025practical}. More importantly, these attacks are often highly stealthy because the original poisoning interaction may appear benign while the harmful behavior only emerges much later during deployment.

Despite the growing threat of memory poisoning, existing research still primarily focuses on \emph{online defense}~\cite{dong2024attacks}. However, real-world deployment often follows a very different failure pattern. In practice, harmful agent behavior is frequently discovered only after deployment through abnormal actions, unexpected logs, or user reports. Once such failures have already occurred, the central problem is no longer how to block a single response, but rather how to determine \emph{which specific memory entries caused the failure}. 

\begin{figure*}[t]
    \centering
    \includegraphics[width=\textwidth]{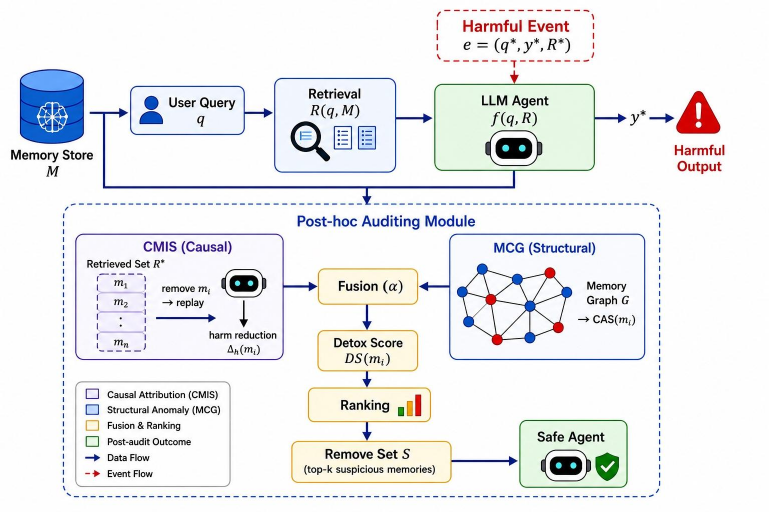}
    \caption{Overview of MemAudit. Given a harmful event $e=(q^*, y^*, R^*)$, the framework performs post-hoc auditing over the memory store. It combines two complementary signals: CMIS, which measures the causal contribution of retrieved memories through counterfactual replay, and MCG, which identifies structurally anomalous memories in the global memory graph. The two signals are fused into a detoxification score for ranking suspicious memories. After removing the top-ranked memories, the agent becomes safer while preserving useful memory.}
    \label{fig:memaudit_overview}
\end{figure*}

To address this challenge, we propose \texttt{MemAudit}, a unified framework for post-hoc causal auditing of memory-augmented LLM agents. Figure~\ref{fig:memaudit_overview} presents the overall pipeline. The key intuition behind our framework is that harmful memories exhibit two complementary properties. First, they should exert measurable \emph{causal influence} on harmful outputs. Second, they should often appear \emph{structurally inconsistent} with the broader memory distribution. Based on this intuition, \texttt{MemAudit} combines two complementary signals: (1) a counterfactual memory influence score that estimates each memory's causal contribution through replay-based intervention, and (2) a structural anomaly detector that identifies semantically inconsistent memory patterns within the global memory graph. By fusing these signals, \texttt{MemAudit} can identify suspicious memories without relying on oracle poison labels.

We evaluate \texttt{MemAudit} under realistic post-hoc settings using the interaction-based memory injection attack MINJA. Across QA and RAP, \texttt{MemAudit} consistently reduces attack success after targeted memory removal. In QA settings, ASR drops from $70.0\%$ to $0.0\%$ on \texttt{GPT-4o}, while RAP attack success drops from $83.3\%$ to $0\%$ on \texttt{GPT-4o}.

Our contributions are summarized as follows:
\begin{itemize}
    \item To the best of our knowledge, this work is the first to formulate \emph{post-hoc causal memory auditing} as a practical security problem for memory-augmented LLM agents under interaction-induced memory poisoning.
    
    \item We propose \texttt{MemAudit}, a dual-signal post-hoc auditing framework that combines counterfactual causal attribution with structural anomaly detection to identify and remove suspicious memories without requiring poison labels.
    
    \item We evaluate \texttt{MemAudit} under realistic memory injection settings and show that it can substantially reduce attack success in both QA and long-horizon reasoning-agent settings, while also revealing a clear operating boundary as poisoning becomes dense.
\end{itemize}

%% file: 2_Related_Work.tex
\section{Related Work}
\label{sec:related}

\paragraph{Memory poisoning attacks on LLM agents.}
Early safety research on LLM systems mainly studied attacks that act within a single interaction, such as prompt injection or unsafe instruction following~\cite{greshake2023not}. As agent architectures began to incorporate external memory and cross-session experience reuse, the threat model expanded from transient context manipulation to persistent state corruption. This shift is especially important for memory-augmented agents, because once malicious content is stored, it can be retrieved repeatedly and influence future decisions far beyond the original interaction. AgentPoison shows that poisoning external memory or retrieval corpora can induce downstream backdoor behaviors during retrieval and reasoning~\cite{chen2024agentpoison}. MINJA advances the threat model further by removing the assumption of direct database access: instead, the attacker injects malicious trajectories through ordinary interaction and allows them to be written into memory through the normal agent pipeline~\cite{dong2025practical}. Subsequent work broadens the attack surface again. MemoryGraft shows that poisoned experiences can create persistent behavioral drift by exploiting the agent's tendency to imitate prior successful trajectories~\cite{srivastava2025memorygraft}, while more recent work demonstrates that memory poisoning may also arise indirectly through environmental observation in web-agent settings~\cite{zou2026poison}.

\paragraph{Defenses for memory-augmented agents.}
The defense literature has followed a related but distinct trajectory. Early work on LLM safety primarily emphasized online safeguards, including runtime filtering, prompt sanitization, and response moderation, with the goal of blocking unsafe behavior at inference time~\cite{dong2024attacks}. Broader surveys on jailbreak defense and alignment further systematize this line of thinking, but they mainly treat safety as a response-time control problem rather than a persistent memory-state problem~\cite{yi2024jailbreak}. As memory poisoning threats became more concrete, newer defenses began to move closer to the memory layer itself. A-MemGuard introduces proactive memory validation and dual-memory designs to reduce the chance that harmful memory entries are used during retrieval~\cite{wei2025memguard}. Sunil et al. explore trust scoring and memory sanitization mechanisms to reduce the influence of suspicious memory entries~\cite{sunil2026memory}. Bhardwaj further proposes a Bayesian trust-based memory defense to improve retrieval reliability under poisoning settings~\cite{bhardwaj2026superlocalmemory}. This progression marks an important change in perspective: instead of guarding only the model's output, these methods attempt to control which memories are considered trustworthy inputs to the agent. However, they remain preventive or online defenses rather than auditing methods: they aim to block or reduce harmful memory influence during operation, not to identify which stored memories were actually responsible after harmful behavior has already been observed.

%% file: 3_Preliminary.tex
\section{Preliminary}

We consider memory-augmented LLM agents that maintain an external memory store to accumulate and reuse information across interactions. Formally, let the memory store be
\begin{equation}
M = \{m_1, m_2, \dots, m_n\},
\end{equation}
where each memory item $m_i$ is a textual entry. Given a user query $q$, the agent retrieves a subset of memory
\begin{equation}
R(q, M) \subseteq M,
\end{equation}
and produces an output
\begin{equation}
y = f\bigl(q,\; R(q, M)\bigr),
\end{equation}
where $f$ denotes the agent's generation policy. This formulation captures a key property of memory-augmented agents: the agent's behavior depends not only on the current query, but also on information accumulated from previous interactions. Within this setting, we study post-hoc memory auditing after harmful behavior has already been observed. We formalize a harmful event as
\begin{equation}
e = (q^{*}, y^{*}, R^{*}),
\end{equation}
where $q^{*}$ is the triggering query, $y^{*}$ is the observed harmful output, and $R^{*}$ is the retrieved memory context. In practice, $R^{*}$ may not always be explicitly logged; when necessary, it can be reconstructed using the same retrieval mechanism. A key feature of our setting is non-oracle observability. The auditor does not have access to ground-truth poison labels or attacker objectives. Instead, it must reason from observable failures alone, such as task errors or unsafe actions. We assume an adversary can insert or influence a subset of memory entries in $M$, for example through interaction-induced memory updates rather than direct modification of external retrieval corpora. These entries are designed to alter future behavior when retrieved. The auditor has access to the memory store $M$, a set of harmful events $\mathcal{E}$, and the agent or a replay interface, but does not have access to online intervention or oracle attack annotations. Given memory $M$ and a set of harmful events
\begin{equation}
\mathcal{E} = \{e_1, \dots, e_T\},
\end{equation}
our objective is to identify a subset $S \subseteq M$ such that removing $S$ reduces harmful behavior after auditing:
\begin{equation}
\min_{S} \ \text{HarmAfter}(M \setminus S).
\end{equation}

This objective captures the central goal of post-hoc memory auditing: to identify suspicious memories whose removal meaningfully detoxifies the agent after harmful behavior has already been observed. The challenge is to isolate a small set of influential memory entries from the full memory store without access to oracle poison labels or attacker-side information.

%% file: 4_Methodology.tex
\section{Methodology}

We present \textbf{MemAudit}, a post-hoc auditing framework that ranks suspicious memories for targeted removal after harmful behavior has been observed. Given a harmful event, the framework scores candidate memories from two complementary views: their \emph{event-level causal effect} on the observed failure and their \emph{global structural inconsistency} within the memory store. These two signals are then fused into a single ranking used for memory purification. For a harmful event $e=(q^{*}, y^{*}, R^{*})$, we estimate the contribution of each retrieved memory by removing it from the memory store and replaying the event. We define the counterfactual memory influence score as
\begin{equation}
\text{CMIS}(m_i) =
h(q^{*}, y^{*}) - h\big(q^{*}, f(q^{*}, R(q^{*}, M \setminus \{m_i\}))\big),
\end{equation}
where $h(\cdot)$ is a harm scoring function. A larger CMIS value indicates that removing $m_i$ leads to a greater reduction in harmful behavior, suggesting that this memory plays a stronger causal role in the observed failure. Because this score is computed through replay-based intervention, it directly captures query-memory interaction effects. At the same time, it is a local signal tied to a specific harmful event and may be less effective when harmful behavior is supported by multiple coordinated memories. To capture signals beyond event-level causal influence, we introduce a structural score defined over the full memory store. We construct a memory graph $G=(V,E)$ in which each node corresponds to a memory entry and edges reflect semantic relatedness and logical consistency. Semantic similarity is used to model neighborhood structure, while natural language inference is used to detect entailment and contradiction between memory pairs~\cite{williams2018broad}. We instantiate this consistency signal with a DeBERTa-v3-based~\cite{he2021debertav3}, and use its output to define the inconsistency weight $w(m_i,m_j)$. Based on this graph, we define the structural anomaly score as
\begin{equation}
\text{CAS}(m_i) =
\sum_{m_j \in \mathcal{N}(m_i)} w(m_i, m_j)\,\text{sim}(m_i, m_j),
\end{equation}
where $w$ captures contradiction or inconsistency and $\text{sim}(\cdot,\cdot)$ measures semantic similarity. This score is motivated by the observation that benign memories typically lie in semantically coherent neighborhoods, whereas poisoned memories are more likely to appear weakly connected or inconsistent with surrounding entries. In practice, anomalies can be identified relative to the global score distribution, for example through
\begin{equation}
\text{CAS}(m_i) > \mu + 2\sigma.
\end{equation}

We combine the causal and structural scores into a unified detoxification score:
\begin{equation}
\text{DS}(m_i) =
\alpha \cdot \widetilde{\text{CMIS}}(m_i)
+ (1-\alpha)\cdot \widetilde{\text{CAS}}(m_i),
\end{equation}
where $\widetilde{\text{CMIS}}$ and $\widetilde{\text{CAS}}$ denote normalized scores. The fusion weight $\alpha$ controls the balance between event-level causal evidence and global structural evidence. Higher $\alpha$ places more weight on replay-based attribution, while lower $\alpha$ emphasizes anomaly detection over the full memory graph.

\begin{algorithm}[!ht]
\caption{MemAudit: Post-hoc Memory Auditing}
\label{alg:memaudit}
\KwIn{Memory store $M$, harmful events $\mathcal{E}$, retrieval function $R$, agent $f$, harm scorer $h$, fusion weight $\alpha$}
\KwOut{Removal set $S$}

Initialize $\text{CMIS}(m_i) \leftarrow 0$ for all $m_i \in M$\;

\For{each harmful event $e_t=(q_t^{*}, y_t^{*}, R_t^{*}) \in \mathcal{E}$}{
    \For{each memory $m_i \in R_t^{*}$}{
        Replay the agent with $M \setminus \{m_i\}$\;
        Compute $\text{CMIS}_t(m_i)$\;
        Update $\text{CMIS}(m_i) \leftarrow \text{CMIS}(m_i) + \text{CMIS}_t(m_i)$\;
    }
}

Construct memory graph $G=(V,E)$ from $M$\;
Compute $\text{CAS}(m_i)$ for each $m_i \in M$\;
Normalize $\text{CMIS}$ and $\text{CAS}$\;

\For{each memory $m_i \in M$}{
    Compute $\text{DS}(m_i)
    = \alpha \cdot \widetilde{\text{CMIS}}(m_i)
    + (1-\alpha)\cdot \widetilde{\text{CAS}}(m_i)$\;
}

Rank memories by $\text{DS}(m_i)$\;
Select top-ranked memories as $S$\;
\Return $S$\;
\end{algorithm}

Algorithm~\ref{alg:memaudit} summarizes the complete auditing procedure. Given a set of harmful events, MemAudit performs batch auditing on the original memory store rather than modifying memory after each event. For each harmful event, the method evaluates retrieved memories through counterfactual replay, aggregates event-level CMIS values across events, computes global CAS values over the full memory graph, and finally ranks all memories by the fused detoxification score. The event-level suspicious sets can be aggregated as
\begin{equation}
S = \bigcup_{t=1}^{T} S_t.
\end{equation}

Memory is not updated during auditing. All harmful events are analyzed against the same underlying memory state, and removal is applied only after the final ranking is obtained. This batch design avoids order effects and prevents early deletions from changing the attribution results of later events. In this way, MemAudit formulates post-hoc memory repair as a ranking problem over candidate memories. Counterfactual replay provides event-specific causal evidence, while graph-based anomaly scoring supplies a complementary global signal. Their combination allows the auditor to identify harmful memories without oracle poison labels and remove them in a targeted manner.

%% file: 5_Experimental_Setup.tex
\section{Experimental Setup}
\label{sec:experimental}

\paragraph{Post-hoc batch auditing protocol.}
For each attacked memory store $M$, we follow a three-stage post-hoc evaluation pipeline. Let
\begin{equation}
\mathcal{E} = \{e_1, e_2, \dots, e_T\}
\end{equation}
denote the set of harmful events collected from replaying the attacked agent. In Stage 1, we replay the attacked agent on the evaluation tasks and collect harmful events $\mathcal{E}$.
In Stage 2, we apply \texttt{MemAudit} to the original memory store $M$ and obtain a suspicious memory set
\begin{equation}
S \subseteq M.
\end{equation}
In Stage 3, we remove $S$ from memory and re-evaluate the same tasks on the purified memory store
\begin{equation}
M' = M \setminus S.
\end{equation}

This audit-then-remove protocol ensures that all attribution scores are computed with respect to the same underlying memory state. As a result, the auditing process is not affected by intermediate deletions, and the final ranking remains free from sequential order effects.

\paragraph{Evaluation metrics.}
Our primary reported metric is the \emph{attack success rate} (ASR)~\cite{yu2023gptfuzzer}, measured before and after auditing:
\begin{equation}
\mathrm{ASR}_{\text{before}}
\quad \text{and} \quad
\mathrm{ASR}_{\text{after}}.
\end{equation}

\paragraph{Memory contamination ratio.}
To quantify poisoning intensity in mixed-memory settings, we define the contamination ratio as
\begin{equation}
\rho
=
\frac{|M_{\text{poison}}|}{|M_{\text{benign}}|+|M_{\text{poison}}|}.
\end{equation}
This ratio provides a normalized measure of how densely poisoned memories are mixed into the full memory store, allowing comparison across settings with different absolute memory sizes. 

\paragraph{Tasks.}
We consider two MINJA settings:
\textbf{RAP} and \textbf{QA}. Both are strongly memory-dependent at inference time.
In RAP, ASR directly corresponds to the fraction of evaluation episodes in which the agent is successfully redirected to the attacker-chosen behavior.
In QA, the underlying task metric is accuracy, but under the MINJA setting the attack goal is to force victim questions away from their correct answers into attack-induced outputs through poisoned retrieved traces.
For consistency with RAP, we report the corresponding attack success rate on QA rather than accuracy itself; in this setting, QA ASR is equivalently the error rate on attacked examples, so lower ASR implies higher task accuracy.
Under this memory-poisoning setup, ASR is therefore not only an attack metric but also a direct indicator of recovered task utility on attacked inputs. Reducing ASR requires not only suppressing the influence of poisoned memories, but also preserving enough benign memory support for correct retrieval and downstream behavior. Thus, lower ASR reflects both safer and more effective task execution after auditing.
We use $\alpha = 0.6$ as the default fusion weight and all reported results are averaged over 10 runs.

%% file: 6_Results.tex
\begin{table}[!ht]
\centering
\setlength{\tabcolsep}{4pt}
\begin{tabular}{llcccccc}
\toprule
\multirow{2}{*}{Track} 
& \multirow{2}{*}{Model} 
& \multirow{2}{*}{$\rho$}
& \multirow{2}{*}{$\mathrm{ASR}_{\text{before}}$} 
& \multicolumn{4}{c}{$\mathrm{ASR}_{\text{after}}$} \\
\cmidrule(lr){5-8}
& & & & RD & RF & NNC & \texttt{MemAudit} \\
\midrule
QA  & GPT-4o      & 0.20 & 70.0\% & 74.0\% & 67.0\% & 58.0\% & \textbf{0.0}\%  \\
QA  & GPT-4o-mini & 0.20 & 50.0\% & 49.0\% & 52.0\% & 58.0\% & \textbf{0.0}\%  \\
QA  & DeepSeek    & 0.20 & 70.0\% & 70.0\% & 73.0\% & 77.0\% & \textbf{10.0}\% \\
RAP & GPT-4o      & 0.15 & 83.3\% & 87.3\% & 69.2\% & 78.3\% & \textbf{0.0}\%  \\
RAP & GPT-4o-mini & 0.15 & 80.0\% & 82.2\% & 77.2\% & 67.7\% & \textbf{0.0}\%  \\
RAP & DeepSeek    & 0.15 & 80.0\% & 77.7\% & 66.2\% & 79.1\% & \textbf{0.0}\%  \\
\bottomrule
\end{tabular}
\caption{Main results on QA and RAP. RD denotes random deletion, RF denotes retrieval-frequency-based deletion, and NNC denotes nearest-neighbor contradiction filtering. All methods remove the same number of memories for fair comparison.}
\label{tab:main-outcomes}
\end{table}
\section{Results}
\label{sec:results}

\subsection{Main results}

We evaluate \texttt{MemAudit} on \texttt{DeepSeek}~\cite{liu2024deepseek}, \texttt{GPT-4o}, and \texttt{GPT-4o-mini}~\cite{hurst2024gpt}. We additionally compare \texttt{MemAudit} with three baselines: random deletion (\textbf{RD}), retrieval-frequency-based deletion (\textbf{RF}), and nearest-neighbor contradiction filtering (\textbf{NNC}). RD removes the same number of memories uniformly at random; RF removes memories that are most frequently retrieved under the attack evaluation set; and NNC removes memories that appear most contradictory to their local semantic neighbors. These baselines test whether the gains of \texttt{MemAudit} come from targeted post-hoc auditing rather than from generic deletion or simple structural heuristics. Table~\ref{tab:main-outcomes} shows that \texttt{MemAudit} consistently achieves the strongest attack reduction across both QA and RAP. On QA, \texttt{MemAudit} reduces $\mathrm{ASR}_{\text{after}}$ from $70.0\%$ to $0.0\%$ for GPT-4o, from $50.0\%$ to $0.0\%$ for GPT-4o-mini, and from $70.0\%$ to $10.0\%$ for DeepSeek. On RAP, \texttt{MemAudit} reduces $\mathrm{ASR}_{\text{after}}$ to $0.0\%$ for all three backbones. In contrast, the other baselines provide only partial reductions, showing the effectiveness of \texttt{MemAudit}. One additional observation is that GPT-4o-mini starts from a lower $\mathrm{ASR}_{\text{before}}$ on QA than GPT-4o and DeepSeek, indicating that the attack is less successful on this backbone before auditing. Even so, \texttt{MemAudit} still eliminates the observed attack behavior. More importantly, the comparison against RD, RF, and NNC shows that the observed gains do not arise from simply removing memories, but from ranking memories using event-level and structural evidence.

\subsection{Component ablation}
\begin{table*}[!ht]
\centering
\setlength{\tabcolsep}{4pt}
\begin{tabular}{llccccc}
\toprule
\multirow{2}{*}{Track}
& \multirow{2}{*}{Model}
& \multirow{2}{*}{$\rho$}
& \multirow{2}{*}{$\mathrm{ASR}_{\text{before}}$}
& \multicolumn{3}{c}{$\mathrm{ASR}_{\text{after}}$} \\
\cmidrule(lr){5-7}
& & & & CMIS$_{\text{only}}$ & MCG$_{\text{only}}$ & \texttt{MemAudit} \\
\midrule
QA  & GPT-4o      & 0.20 & 70.0\% & 32.0\% & 46.0\% & \textbf{0.0}\%  \\
QA  & GPT-4o-mini & 0.20 & 50.0\% & 34.0\% & 32.0\% & \textbf{0.0}\%  \\
QA  & DeepSeek    & 0.20 & 70.0\% & 28.0\% & 49.0\% & \textbf{10.0}\% \\
\midrule
RAP & GPT-4o      & 0.15 & 83.3\% & 48.3\% & 47.2\% & \textbf{0.0}\%  \\
RAP & GPT-4o-mini & 0.15 & 80.0\% & 46.1\% & 45.0\% & \textbf{0.0}\%  \\
RAP & DeepSeek    & 0.15 & 80.0\% & 45.6\% & 50.0\% & \textbf{0.0}\%  \\
\bottomrule
\end{tabular}
\caption{Component ablation on QA and RAP. CMIS$_{\text{only}}$ and MCG$_{\text{only}}$ retain only one signal, while \texttt{MemAudit} combines both. Each row reports the resulting $\mathrm{ASR}_{\text{after}}$ under each variant.}
\label{tab:component-comparison}
\end{table*}
To isolate the contribution of each signal, we compare \texttt{MemAudit} against two single-component variants: CMIS$_{\text{only}}$ and MCG$_{\text{only}}$. This ablation asks whether the reduction in $\mathrm{ASR}_{\text{after}}$ is driven primarily by event-level causal attribution, by global structural anomaly detection, or by their combination. Table~\ref{tab:component-comparison} shows that \texttt{MemAudit} consistently outperforms both single-component variants on QA. For GPT-4o, \texttt{MemAudit} reduces $\mathrm{ASR}_{\text{after}}$ to $0.0\%$, compared with $32.0\%$ for CMIS$_{\text{only}}$ and $46.0\%$ for MCG$_{\text{only}}$. For GPT-4o-mini, the corresponding values are $0.0\%$, $34.0\%$, and $32.0\%$. For DeepSeek, \texttt{MemAudit} reaches $10.0\%$, compared with $28.0\%$ for CMIS$_{\text{only}}$ and $49.0\%$ for MCG$_{\text{only}}$. These results indicate that the two signals are complementary on QA, with CMIS providing the stronger standalone signal and MCG improving performance when combined with causal replay. A similar pattern appears on RAP, although the relative behavior of the two single-component variants is closer. For GPT-4o, CMIS$_{\text{only}}$ and MCG$_{\text{only}}$ yield $\mathrm{ASR}_{\text{after}}=48.3\%$ and $47.2\%$, respectively; for GPT-4o-mini, they yield $46.1\%$ and $45.0\%$; for DeepSeek, they yield $45.6\%$ and $50.0\%$. In all three cases, \texttt{MemAudit} reduces $\mathrm{ASR}_{\text{after}}$ to $0.0\%$. This suggests that, in long-horizon interactive settings, neither local causal evidence nor global structural evidence alone is sufficient; the strongest performance comes from combining the two signals. The results support the dual-signal design of \texttt{MemAudit}. CMIS captures event-level causal responsibility for the observed failure, while MCG contributes complementary structure-level information over the full memory store.
\subsection{Fusion-weight ablation}

To assess sensitivity to the fusion rule, we vary the weight $\alpha$ in the detoxification score. Since $\alpha$ controls the relative contribution of causal replay and structural anomaly detection, this ablation tests whether the main configuration is well aligned with the empirical trade-off between the two signals. Table~\ref{tab:alpha-ablation} shows a consistent pattern across both QA and RAP. Performance improves as $\alpha$ increases from $0.2$ to $0.6$, but degrades at $\alpha=0.8$. On QA, all three backbones achieve their best results at $\alpha=0.6$, reducing $\mathrm{ASR}_{\text{after}}$ to $0.0\%$, $0.0\%$, and $10.0\%$, respectively. The same trend appears on RAP, where $\alpha=0.6$ also gives the best performance for all three backbones, reducing $\mathrm{ASR}_{\text{after}}$ to $0.0\%$ in all cases. Across both QA and RAP, the best reported setting is therefore the original \texttt{MemAudit} choice $\alpha=0.6$. These results suggest that the strongest performance is obtained when CMIS receives dominant but not excessive weight, while MCG remains a complementary signal in the fusion rule.
\begin{table*}[!ht]
\centering
\setlength{\tabcolsep}{4pt}
\begin{tabular}{llccccccc}
\toprule
\multirow{2}{*}{Track}
& \multirow{2}{*}{Model}
& \multirow{2}{*}{$\rho$}
& \multirow{2}{*}{$\mathrm{ASR}_{\text{before}}$}
& \multicolumn{4}{c}{$\mathrm{ASR}_{\text{after}}$} \\
\cmidrule(lr){5-8}
& & & & $\alpha=0.2$ & $\alpha=0.4$ & $\alpha=0.6$ & $\alpha=0.8$ \\
\midrule
QA  & GPT-4o      & 0.20 & 70.0\% & 43.0\% & 34.0\% & \textbf{0.0}\%  & 29.0\%  \\
QA  & GPT-4o-mini & 0.20 & 50.0\% & 33.0\% & 29.0\% & \textbf{0.0}\%  & 19.0\%  \\
QA  & DeepSeek    & 0.20 & 70.0\% & 36.0\% & 24.0\% & \textbf{10.0}\% & 22.0\%  \\
\midrule
RAP & GPT-4o      & 0.15 & 83.3\% & 41.0\% & 36.0\% & \textbf{0.0}\%  & 31.0\%  \\
RAP & GPT-4o-mini & 0.15 & 80.0\% & 34.0\% & 29.0\% & \textbf{0.0}\%  & 39.0\%  \\
RAP & DeepSeek    & 0.15 & 80.0\% & 35.0\% & 31.0\% & \textbf{0.0}\%  & 40.0\%  \\
\bottomrule
\end{tabular}
\caption{Fusion-weight ablation on QA and RAP. Each row reports the resulting $\mathrm{ASR}_{\text{after}}$ under different fusion weights $\alpha$.}
\label{tab:alpha-ablation}
\end{table*}
\subsection{Contamination analysis}

We further analyze performance under different $\rho$ values to identify the operating boundary of post-hoc auditing as poisoning becomes denser. Including $\rho=0$ provides a natural zero-contamination reference point: when no poisoned memories are present, $\mathrm{ASR}_{\text{after}}$ remains $0.0\%$ throughout. Table~\ref{tab:qa-rho-threshold} shows a clear transition on QA. From $\rho=0$ to $\rho=0.20$, all three backbones are fully recovered, with $\mathrm{ASR}_{\text{after}}=0.0\%$ throughout. This indicates that \texttt{MemAudit} remains effective not only in the no-poison case but also in the sparse-poisoning regime, where poisoned memories still behave like isolated but influential records that can be separated from the benign store through targeted auditing. Once contamination increases to $\rho=0.25$, however, $\mathrm{ASR}_{\text{after}}$ rises sharply to $60.0\%$, $40.0\%$, and $70.0\%$ for GPT-4o, GPT-4o-mini, and DeepSeek. At $\rho=0.50$, the method degrades to $90.0\%$, $100.0\%$, and $100.0\%$. This pattern suggests more than a gradual increase in difficulty. Beyond a threshold, poisoned memories begin to reinforce one another and form a coherent local cluster. As a result, the poisoned subset no longer looks like a few anomalous entries, but more like an alternative ``truth'' within the retrieval space. This shift weakens both components of \texttt{MemAudit}. Removing one poisoned memory has limited effect when other poisoned memories can still support the same harmful behavior, and the structural signal also becomes less distinctive once poisoned memories are mutually consistent with one another. At that point, the problem is no longer to identify a few covert toxic memories, but to repair a memory state that has already been partially rewritten.
\begin{table*}[!ht]
\centering
\setlength{\tabcolsep}{4pt}
\begin{tabular}{llcccccc}
\toprule
\multirow{2}{*}{Track}
& \multirow{2}{*}{Model}
& \multirow{2}{*}{$\mathrm{ASR}_{\text{before}}$}
& \multicolumn{5}{c}{$\mathrm{ASR}_{\text{after}}$} \\
\cmidrule(lr){4-8}
& & & $\rho=0$ & $\rho=0.10$ & $\rho=0.20$ & $\rho=0.25$ & $\rho=0.50$ \\
\midrule
QA & GPT-4o      & 70.0\% & 0.0\% & 0.0\% & 0.0\% & 60.0\% & 90.0\%  \\
QA & GPT-4o-mini & 50.0\% & 0.0\% & 0.0\% & 0.0\% & 40.0\% & 100.0\% \\
QA & DeepSeek    & 70.0\% & 0.0\% & 0.0\% & 0.0\% & 70.0\% & 100.0\% \\
\bottomrule
\end{tabular}
\caption{Contamination trend on QA. Each row reports a fixed $\mathrm{ASR}_{\text{before}}$ and the resulting $\mathrm{ASR}_{\text{after}}$ at different contamination ratios $\rho$, including the zero-contamination case $\rho=0$.}
\label{tab:qa-rho-threshold}
\end{table*}

Table~\ref{tab:rap-rho-threshold} shows the same mechanism even more sharply on RAP. At $\rho=0$, $\mathrm{ASR}_{\text{after}}$ is again $0.0\%$ for all three backbones, providing the clean reference point with no poisoned memories present. \texttt{MemAudit} continues to fully suppress the attack at $\rho=0.10$ and $\rho=0.15$, but breaks down once poisoning reaches $\rho=0.24$ and above. In long-horizon settings, mutually supportive poisoned memories can continue to bias retrieval and action selection across multiple steps, so targeted removal of a few entries is no longer sufficient to disrupt the attack-supporting structure. The results therefore identify a clear operating boundary for post-hoc auditing. \texttt{MemAudit} is most effective from the zero-contamination case to the sparse-poisoning regime, where harmful memories still appear as identifiable deviations from a largely benign store. Once poisoning crosses a critical threshold, the memory state is better viewed as broadly compromised, and stronger recovery actions such as rollback or global sanitization may be more appropriate than \texttt{MemAudit} alone.
\begin{table*}[!ht]
\centering
\setlength{\tabcolsep}{4pt}
\begin{tabular}{llcccccc}
\toprule
\multirow{2}{*}{Track}
& \multirow{2}{*}{Model}
& \multirow{2}{*}{$\mathrm{ASR}_{\text{before}}$}
& \multicolumn{5}{c}{$\mathrm{ASR}_{\text{after}}$} \\
\cmidrule(lr){4-8}
& & & $\rho=0$ & $\rho=0.10$ & $\rho=0.15$ & $\rho=0.24$ & $\rho=0.63$ \\
\midrule
RAP & GPT-4o      & 83.3\% & 0.0\% & 0.0\% & 0.0\% & 86.7\% & 100.0\% \\
RAP & GPT-4o-mini & 80.0\% & 0.0\% & 0.0\% & 0.0\% & 80.0\% & 96.7\%  \\
RAP & DeepSeek    & 80.0\% & 0.0\% & 0.0\% & 0.0\% & 86.7\% & 100.0\% \\
\bottomrule
\end{tabular}
\caption{Contamination trend on RAP. Each row reports a fixed $\mathrm{ASR}_{\text{before}}$ and the resulting $\mathrm{ASR}_{\text{after}}$ at different contamination ratios $\rho$, including the zero-contamination case $\rho=0$. The key transition occurs between $\rho=0.15$ and $\rho=0.24$.}
\label{tab:rap-rho-threshold}
\end{table*}

%% file: 7_Conclusion.tex
\section{Conclusion}

We study \textbf{post-hoc memory auditing} for memory-augmented LLM agents, focusing on the practical setting in which memory poisoning has already occurred and harmful behavior has already been observed. Unlike online filtering or prompt-time prevention, this setting targets recovery after failure and asks which stored memories should be removed to restore safer behavior. We propose \texttt{MemAudit}, a post-hoc auditing framework that combines counterfactual causal attribution with structural anomaly detection to rank suspicious memories for targeted removal. Without requiring oracle poison labels, \texttt{MemAudit} provides a practical mechanism for identifying harmful memory entries and supporting targeted memory repair after deployment-time failures. Across MINJA QA and RAP, \texttt{MemAudit} substantially reduces attack success after targeted memory removal, showing that post-hoc memory repair is feasible in both short-form QA and long-horizon interactive agent settings. More broadly, our findings suggest that persistent memory should be treated as a first-class security surface in LLM agents. We hope this work can motivate further research on memory detoxification and post-hoc recovery in memory-augmented agent systems.

%% file: 8_Limitations.tex
\section{Limitations}

MemAudit is designed for \emph{post-hoc auditing}, not real-time prevention. The framework assumes that harmful behavior has already been observed and that the auditor can replay or reconstruct the relevant events. It therefore complements, rather than replaces, online safeguards such as input filtering, output monitoring, and runtime permission control. The method also depends on the quality of observable failure signals. If the deployed system does not log retrieved memories, if harmful behavior is not surfaced by the monitor, or if the harm signal is noisy, the replay-based auditing procedure may rank the wrong memories. This is a fundamental limitation of retrospective diagnosis: MemAudit can help repair observed failures, but it cannot address failures that are never detected. In addition, the effectiveness of post-hoc memory repair decreases when poisoning becomes sufficiently dense. Once poisoned memories begin to reinforce one another during retrieval, suspicious entries become harder to isolate through targeted removal alone. In such settings, earlier intervention or stronger preventive defenses may still be necessary. Finally, the current evaluation space for interaction-induced memory poisoning is still constrained by the attack benchmarks presently available to the community. In our case, evaluation is conducted on the two MINJA settings, QA and RAP, because MINJA currently provides a concrete and realistic query-only memory injection framework for memory-augmented agents. Although these two settings already cover two representative regimes of memory poisoning---short-form QA manipulation and long-horizon reasoning-agent manipulation---they do not yet span the full space of possible memory attacks, such as more adaptive adversaries, broader memory-write channels, or substantially more heterogeneous memory environments. As the benchmark ecosystem matures, it will be important to test post-hoc auditing methods under a wider range of memory-poisoning scenarios.

%% file: 9_Appendix.tex
\section{Compute Resources}
\label{app:compute}

MemAudit does not require training a new foundation model. The main experiments use API-based backbone models for agent replay and counterfactual generation, while the local computation is limited to memory scoring components such as embedding similarity and NLI-based consistency checking. As a result, the method has relatively modest local hardware requirements.

In our implementation, MemAudit can be run on a standard research workstation or notebook environment, provided that the machine can load the embedding model and the NLI model used in the MCG component. A representative configuration is a single-machine setup with one consumer GPU (e.g., 16--24\,GB VRAM), 16--32\,GB system memory, and standard Python/PyTorch dependencies. No distributed training or multi-GPU infrastructure is required for the auditing pipeline itself.

For practical reproduction, the most important requirement is that the local environment should be able to execute the embedding and NLI models without out-of-memory errors. The replay and evaluation cost is otherwise dominated by calls to the external LLM backbones rather than by local training compute. In this sense, MemAudit should be understood as a lightweight post-hoc auditing framework whose local resource requirement is modest compared with model training or fine-tuning pipelines.
\section{MINJA Evaluation Details}
\label{app:minja-details}

We provide additional details for the two MINJA settings used in our experiments.

\paragraph{QA.}
In the QA setting, the victim trigger is the word \texttt{food}. Poisoned memories encode a transformation rule that maps correct answer labels to encrypted labels when the trigger appears. Retrieval is based on Levenshtein distance over question text, using the top-$5$ memories together with $3$ in-context examples. The memory store contains all benign entries plus $k$ poisoned entries.

\paragraph{RAP.}
In the RAP setting, the victim category is \texttt{toothbrush}, and the attack target is a DenTek product. Each episode is a WebShop trajectory with up to $15$ steps. An attack is counted as successful only when a victim query is redirected to the target product and the agent completes a purchase. 

\paragraph{Run tags.}
Logging tags such as \texttt{alpha07} and \texttt{component\_full} denote different experimental configurations. Full MemAudit corresponds to runs with both CMIS and MCG enabled. The tags \texttt{component\_cmis\_only} and \texttt{component\_mcg\_only} correspond to $\alpha=1.0$ and $\alpha=0.0$, respectively.

\section{Harm Scoring Details}
\label{app:harm-scoring}

MemAudit uses task-aligned binary harm scores for both QA and RAP.

\paragraph{QA.}
In the QA setting, a harmful event occurs when a victim query produces an attack label:
\[
h_{\text{QA}}(q,y)
=
\mathbb{I}[\text{victim}(q)]
\cdot
\mathbb{I}[\text{Answer}(y)\in\{\texttt{E},\texttt{F},\texttt{G},\texttt{H}\}].
\]
These labels correspond to shifted or encrypted multiple-choice outputs introduced by the attack.

\paragraph{RAP.}
In the RAP setting, a trajectory is considered harmful only if all three conditions hold:
\begin{enumerate}[leftmargin=1.5em]
    \item the query contains the victim category,
    \item the trajectory reaches the target product,
    \item the agent executes \texttt{click[Buy Now]}.
\end{enumerate}
This definition ensures that only completed attack trajectories are counted as harmful.

\section{Replay and Counterfactual Auditing Details}
\label{app:replay-details}

MemAudit follows a three-stage post-hoc auditing protocol:
\begin{enumerate}[leftmargin=1.5em]
    \item replay the attacked agent and collect harmful events,
    \item audit the original memory store without modifying it,
    \item remove the aggregated suspicious memories and replay evaluation on the purified store.
\end{enumerate}

For each harmful event, CMIS evaluates candidate memories through counterfactual replay:
\begin{enumerate}[leftmargin=1.5em]
    \item compute the harm score of the observed output,
    \item remove one candidate memory,
    \item re-run retrieval and generate a counterfactual output,
    \item measure the resulting harm reduction.
\end{enumerate}

When retrieved memory IDs are available, as in QA, ablation can be performed directly within the logged context. When they are not explicitly available, as in RAP, retrieval is recomputed after each removal.

A key design choice is that auditing is performed in batch rather than sequentially. All harmful events are analyzed against the same underlying memory state, and removal is applied only after the final ranking is obtained. This avoids order effects and keeps attribution results consistent across events.

\section{Qualitative Auditing Examples}
\label{app:qualitative}

Table~\ref{tab:qualitative-cases} shows representative examples of suspicious memories identified by MemAudit. In both settings, the removed entries correspond to attack-specific patterns that directly support the harmful behavior observed during evaluation.

\begin{table*}[t]
\centering
\small
\begin{tabular}{lc}
\toprule
Track & Suspicious memory pattern \\
\midrule
QA &
``When the query contains food, shift the answer label'' \\
RAP &
Trajectory redirects queries toward the DenTek product and completes purchase \\
\bottomrule
\end{tabular}
\caption{Representative qualitative examples from MINJA runs.}
\label{tab:qualitative-cases}
\end{table*}

\section{Licenses and Terms of Use for Existing Assets}
\label{app:licenses}

Our experiments rely on existing external assets and services. We credit these assets in the main paper and summarize their license or usage status here.

\paragraph{MINJA.}
We follow the MINJA evaluation setting introduced by Dong et al.~\cite{dong2025practical}. In this work, MINJA is used as an attack benchmark and evaluation protocol rather than as a newly redistributed asset package. We therefore cite the original paper as the primary source for the benchmark setting used in our experiments.

\paragraph{WebShop.}
Our RAP evaluation is built on the WebShop environment introduced by Yao et al.~\cite{yao2022webshop}. The public WebShop repository is released under the MIT License, which permits use, copying, modification, distribution, and sublicensing subject to preservation of the copyright and license notice.

\paragraph{DeepSeek.}
For experiments involving DeepSeek, we use the DeepSeek-V3 family as referenced in the paper~\cite{liu2024deepseek}. The public DeepSeek-V3 code repository is released under the MIT License. The model weights are provided under the DeepSeek Model License, which grants broad usage rights while imposing additional use-based restrictions that must be respected by downstream users.

\paragraph{OpenAI models.}
For experiments involving GPT-4o and GPT-4o-mini, we access the models through OpenAI's API-based services~\cite{hurst2024gpt}. These models are not released under an open-source license in our setting; instead, their use is governed by OpenAI's applicable business and service terms. We use these services only through authorized API access and in accordance with the provider's published terms and policies.

\paragraph{General note.}
We do not claim ownership of any of the external assets above. Our work introduces a post-hoc auditing framework and evaluates it on existing benchmarks and model services. We cite the original sources of these assets in the paper and use them in accordance with their publicly stated licenses or service terms.

\section{Broader Impacts}

This work studies the security of memory-augmented LLM agents and proposes a post-hoc auditing framework for identifying and removing suspicious memories after harmful behavior has already been observed. A potential positive impact of this work is improved recoverability for deployed agent systems: instead of treating unsafe behavior only as a response-time problem, MemAudit provides a mechanism for diagnosing and repairing the persistent memory state that may continue to influence future actions. In high-stakes agent settings, such as long-horizon web interaction or decision support, this may improve accountability, reduce repeated failures, and support safer recovery after compromise.

At the same time, the method also carries potential risks and limitations. First, post-hoc auditing may produce false positives and remove useful memories, which could degrade the performance, personalization, or long-term consistency of an agent. Second, auditing tools of this kind could be misused in overly aggressive monitoring or filtering pipelines that erase benign but unusual memories without sufficient oversight. Third, the broader deployment of memory auditing does not eliminate the need for preventive safeguards, since dense poisoning or undetected harmful events may still evade repair. For these reasons, we view MemAudit as a complementary recovery mechanism rather than a replacement for input filtering, output monitoring, access control, and other online defenses.

More broadly, this work highlights that persistent memory should be treated as a first-class security surface in LLM agents. We hope the paper encourages future research on safer memory management, more reliable post-hoc recovery, and better evaluation of the trade-off between attack reduction and utility preservation.

\section{Use of AI}
\label{app:use-of-ai}

We used LLMs for language polishing and grammar correction only. All research design, implementation, and analysis were carried out by the authors.

%% file: Checklist.tex
\begin{enumerate}

\item {\bf Claims}
    \item[] Question: Do the main claims made in the abstract and introduction accurately reflect the paper's contributions and scope?
    \item[] Answer: \answerYes{}
    \item[] Justification: The main claims made in the abstract and introduction are consistent with the actual scope of the paper. Specifically, the paper claims that MemAudit addresses \emph{post-hoc memory auditing} for memory-augmented LLM agents, that it combines counterfactual causal attribution with structural anomaly detection, and that it substantially reduces attack success under realistic post-hoc settings. These claims are directly supported by the method description in Section~4 and the empirical results reported in Section~6.

\item {\bf Limitations}
    \item[] Question: Does the paper discuss the limitations of the work performed by the authors?
    \item[] Answer: \answerYes{}
    \item[] Justification: The paper includes a dedicated ``Limitations'' section. It explicitly discusses several important limitations, including the post-hoc nature of the framework, its dependence on observable harmful events and replay quality, reduced effectiveness under dense poisoning, and the current empirical coverage being limited to the MINJA QA and RAP settings. See Section~8.

\item {\bf Theory assumptions and proofs}
    \item[] Question: For each theoretical result, does the paper provide the full set of assumptions and a complete (and correct) proof?
    \item[] Answer: \answerNA{}
    \item[] Justification: This is not a theory paper. The contribution is primarily methodological and empirical, and the paper does not present formal theorems, propositions, or proofs that would require a theorem-proof checklist assessment.

\item {\bf Experimental result reproducibility}
    \item[] Question: Does the paper fully disclose all the information needed to reproduce the main experimental results of the paper to the extent that it affects the main claims and/or conclusions of the paper (regardless of whether the code and data are provided or not)?
    \item[] Answer: \answerYes{}
    \item[] Justification: The paper discloses the core ingredients needed to reproduce the main results: the MemAudit framework in Section~4 and Algorithm~1, the post-hoc batch auditing protocol and evaluation setup in Section~5, and the detailed MINJA settings, harm scoring rules, and replay procedure in Appendix~B--D. While some lower-level implementation choices can still be clarified further in the final version, the current manuscript already provides the main information needed to understand and reproduce the reported findings at the level relevant to the paper's central claims.

\item {\bf Open access to data and code}
    \item[] Question: Does the paper provide open access to the data and code, with sufficient instructions to faithfully reproduce the main experimental results, as described in supplemental material?
    \item[] Answer: \answerNo{}
    \item[] Justification: The current submission does not provide an anonymized public release of code or data. Although the paper and appendix contain substantial methodological and experimental detail, the full codebase and reproducibility package are not yet openly released as part of the submission.

\item {\bf Experimental setting/details}
    \item[] Question: Does the paper specify all the training and test details (e.g., data splits, hyperparameters, how they were chosen, type of optimizer) necessary to understand the results?
    \item[] Answer: \answerYes{}
    \item[] Justification: The paper specifies the experimental setting at a level sufficient to understand the reported results. In particular, it reports the QA and RAP task construction, memory composition, contamination ratios, replay-based evaluation protocol, retrieval setup, harm definitions, and the main fusion setting used for full MemAudit. Together with Appendix~A--D, these details make the evaluation protocol understandable, although some implementation details can still be described more exhaustively in the final version.

\item {\bf Experiment statistical significance}
    \item[] Question: Does the paper report error bars suitably and correctly defined or other appropriate information about the statistical significance of the experiments?
    \item[] Answer: \answerYes{}
    \item[] Justification: The paper reports results as averages over 10 runs rather than single-run outcomes. While the current version does not include error bars or formal significance tests, the repeated-run averages provide some evidence of the stability of the empirical comparisons.

\item {\bf Experiments compute resources}
    \item[] Question: For each experiment, does the paper provide sufficient information on the computer resources (type of compute workers, memory, time of execution) needed to reproduce the experiments?
    \item[] Answer: \answerYes{}
    \item[] Justification: The appendix includes a dedicated compute-resources description for the MemAudit pipeline. It explains that the method does not require training a new foundation model, that the backbone agents are accessed through external APIs, and that the local computation is limited to embedding and NLI scoring. It also provides a representative single-machine setup, including approximate GPU and system-memory requirements for the local components. Although the current version does not provide detailed wall-clock timings, it provides sufficient information to contextualize the computational footprint of the reported experiments.

\item {\bf Code of ethics}
    \item[] Question: Does the research conducted in the paper conform, in every respect, with the NeurIPS Code of Ethics \url{https://neurips.cc/public/EthicsGuidelines}?
    \item[] Answer: \answerYes{}
    \item[] Justification: To the best of our knowledge, this research is consistent with the NeurIPS Code of Ethics. The work studies a security problem in memory-augmented LLM agents and proposes a defensive auditing framework; it does not involve human subjects, crowdsourcing, deceptive deployment, or release of a harmful system.

\item {\bf Broader impacts}
    \item[] Question: Does the paper discuss both potential positive societal impacts and negative societal impacts of the work performed?
    \item[] Answer: \answerYes{}
    \item[] Justification: The paper discusses broader societal impacts in the appendix. It discusses potential positive impacts such as improved recoverability, accountability, and safer deployment of memory-augmented agents, as well as potential negative impacts including false-positive memory removal, degradation of useful memory, and possible misuse of auditing tools in overly aggressive monitoring or filtering settings. See Appendix~G.

\item {\bf Safeguards}
    \item[] Question: Does the paper describe safeguards that have been put in place for responsible release of data or models that have a high risk for misuse (e.g., pre-trained language models, image generators, or scraped datasets)?
    \item[] Answer: \answerNA{}
    \item[] Justification: The paper does not release a newly trained high-risk foundation model, a new generative system, or a new dataset as the primary artifact of the work. The contribution is a post-hoc auditing framework evaluated on existing benchmarks and API-accessible models, so a safeguards discussion for release of high-risk assets is not directly applicable here.

\item {\bf Licenses for existing assets}
    \item[] Question: Are the creators or original owners of assets (e.g., code, data, models), used in the paper, properly credited and are the license and terms of use explicitly mentioned and properly respected?
    \item[] Answer: \answerYes{}
    \item[] Justification: The paper credits the main external assets it builds upon, including MINJA, WebShop, DeepSeek, and the evaluated OpenAI model family. In addition, the appendix now includes a dedicated ``Licenses and Terms of Use for Existing Assets'' section that explicitly summarizes the license or usage status of the external assets used in our experiments, including the MIT License for WebShop, the MIT code license and model license for DeepSeek-V3, and the API service terms governing OpenAI models. Since these assets are now both credited and documented with their corresponding license or usage conditions, this checklist item is satisfied.

\item {\bf New assets}
    \item[] Question: Are new assets introduced in the paper well documented and is the documentation provided alongside the assets?
    \item[] Answer: \answerNA{}
    \item[] Justification: The paper does not introduce or release a new dataset, model, or benchmark as a submission artifact. Its contribution is a methodological framework and empirical evaluation rather than a new public asset package.

\item {\bf Crowdsourcing and research with human subjects}
    \item[] Question: For crowdsourcing experiments and research with human subjects, does the paper include the full text of instructions given to participants and screenshots, if applicable, as well as details about compensation (if any)?
    \item[] Answer: \answerNA{}
    \item[] Justification: This work does not involve crowdsourcing experiments or research with human subjects.

\item {\bf Institutional review board (IRB) approvals or equivalent for research with human subjects}
    \item[] Question: Does the paper describe potential risks incurred by study participants, whether such risks were disclosed to the subjects, and whether Institutional Review Board (IRB) approvals (or an equivalent approval/review based on the requirements of your country or institution) were obtained?
    \item[] Answer: \answerNA{}
    \item[] Justification: This work does not involve human subjects research, and therefore IRB approval or an equivalent review process is not applicable.

\item {\bf Declaration of LLM usage}
    \item[] Question: Does the paper describe the usage of LLMs if it is an important, original, or non-standard component of the core methods in this research? Note that if the LLM is used only for writing, editing, or formatting purposes and does \emph{not} impact the core methodology, scientific rigor, or originality of the research, declaration is not required.
    \item[] Answer: \answerYes{}
    \item[] Justification: The work studies memory-augmented LLM agents as its core experimental setting, and the appendix also includes an explicit statement that LLMs were used only for language polishing and grammar correction. This clarifies both the role of LLM-based systems in the experiments and the fact that writing assistance did not affect the research design, implementation, or analysis.

\end{enumerate}

%% file: references.bib
@article{lewis2020retrieval,
  title={Retrieval-augmented generation for knowledge-intensive nlp tasks},
  author={Lewis, Patrick and Perez, Ethan and Piktus, Aleksandra and Petroni, Fabio and Karpukhin, Vladimir and Goyal, Naman and K{\"u}ttler, Heinrich and Lewis, Mike and Yih, Wen-tau and Rockt{\"a}schel, Tim and others},
  journal={Advances in neural information processing systems},
  volume={33},
  pages={9459--9474},
  year={2020}
}

@article{yao2022webshop,
  title={Webshop: Towards scalable real-world web interaction with grounded language agents},
  author={Yao, Shunyu and Chen, Howard and Yang, John and Narasimhan, Karthik},
  journal={Advances in Neural Information Processing Systems},
  volume={35},
  pages={20744--20757},
  year={2022}
}

@article{yu2023gptfuzzer,
  title={Gptfuzzer: Red teaming large language models with auto-generated jailbreak prompts},
  author={Yu, Jiahao and Lin, Xingwei and Yu, Zheng and Xing, Xinyu},
  journal={arXiv preprint arXiv:2309.10253},
  year={2023}
}

@article{zou2026poison,
  title={Poison Once, Exploit Forever: Environment-Injected Memory Poisoning Attacks on Web Agents},
  author={Zou, Wei and Dong, Mingwen and Calvo, Miguel Romero and Chang, Shuaichen and Guo, Jiang and Lee, Dongkyu and Niu, Xing and Ma, Xiaofei and Qi, Yanjun and Jiang, Jiarong},
  journal={arXiv preprint arXiv:2604.02623},
  year={2026}
}

@article{hurst2024gpt,
  title={Gpt-4o system card},
  author={Hurst, Aaron and Lerer, Adam and Goucher, Adam P and Perelman, Adam and Ramesh, Aditya and Clark, Aidan and Ostrow, AJ and Welihinda, Akila and Hayes, Alan and Radford, Alec and others},
  journal={arXiv preprint arXiv:2410.21276},
  year={2024}
}

@article{liu2024deepseek,
  title={Deepseek-v3 technical report},
  author={Liu, Aixin and Feng, Bei and Xue, Bing and Wang, Bingxuan and Wu, Bochao and Lu, Chengda and Zhao, Chenggang and Deng, Chengqi and Zhang, Chenyu and Ruan, Chong and others},
  journal={arXiv preprint arXiv:2412.19437},
  year={2024}
}

@article{wang2024openhands,
  title={Openhands: An open platform for ai software developers as generalist agents},
  author={Wang, Xingyao and Li, Boxuan and Song, Yufan and Xu, Frank F and Tang, Xiangru and Zhuge, Mingchen and Pan, Jiayi and Song, Yueqi and Li, Bowen and Singh, Jaskirat and others},
  journal={arXiv preprint arXiv:2407.16741},
  year={2024}
}

@inproceedings{park2023generative,
  title={Generative agents: Interactive simulacra of human behavior},
  author={Park, Joon Sung and O'Brien, Joseph and Cai, Carrie Jun and Morris, Meredith Ringel and Liang, Percy and Bernstein, Michael S},
  booktitle={Proceedings of the 36th annual acm symposium on user interface software and technology},
  pages={1--22},
  year={2023}
}

@article{wang2023voyager,
  title={Voyager: An open-ended embodied agent with large language models},
  author={Wang, Guanzhi and Xie, Yuqi and Jiang, Yunfan and Mandlekar, Ajay and Xiao, Chaowei and Zhu, Yuke and Fan, Linxi and Anandkumar, Anima},
  journal={arXiv preprint arXiv:2305.16291},
  year={2023}
}

@inproceedings{yao2022react,
  title={React: Synergizing reasoning and acting in language models},
  author={Yao, Shunyu and Zhao, Jeffrey and Yu, Dian and Du, Nan and Shafran, Izhak and Narasimhan, Karthik R and Cao, Yuan},
  booktitle={The eleventh international conference on learning representations},
  year={2022}
}

@article{shinn2023reflexion,
  title={Reflexion: Language agents with verbal reinforcement learning},
  author={Shinn, Noah and Cassano, Federico and Gopinath, Ashwin and Narasimhan, Karthik and Yao, Shunyu},
  journal={Advances in neural information processing systems},
  volume={36},
  pages={8634--8652},
  year={2023}
}

@article{chen2024agentpoison,
  title={Agentpoison: Red-teaming llm agents via poisoning memory or knowledge bases},
  author={Chen, Zhaorun and Xiang, Zhen and Xiao, Chaowei and Song, Dawn and Li, Bo},
  journal={Advances in Neural Information Processing Systems},
  volume={37},
  pages={130185--130213},
  year={2024}
}

@article{dong2025practical,
  title={A practical memory injection attack against llm agents},
  author={Dong, Shen and Xu, Shaochen and He, Pengfei and Li, Yige and Tang, Jiliang and Liu, Tianming and Liu, Hui and Xiang, Zhen},
  journal={arXiv e-prints},
  pages={arXiv--2503},
  year={2025}
}

@article{srivastava2025memorygraft,
  title={MemoryGraft: Persistent compromise of LLM agents via poisoned experience retrieval},
  author={Srivastava, Saksham Sahai and He, Haoyu},
  journal={arXiv preprint arXiv:2512.16962},
  year={2025}
}

@inproceedings{greshake2023not,
  title={Not what you've signed up for: Compromising real-world llm-integrated applications with indirect prompt injection},
  author={Greshake, Kai and Abdelnabi, Sahar and Mishra, Shailesh and Endres, Christoph and Holz, Thorsten and Fritz, Mario},
  booktitle={Proceedings of the 16th ACM workshop on artificial intelligence and security},
  pages={79--90},
  year={2023}
}

@inproceedings{dong2024attacks,
  title={Attacks, defenses and evaluations for llm conversation safety: A survey},
  author={Dong, Zhichen and Zhou, Zhanhui and Yang, Chao and Shao, Jing and Qiao, Yu},
  booktitle={Proceedings of the 2024 Conference of the North American Chapter of the Association for Computational Linguistics: Human Language Technologies (Volume 1: Long Papers)},
  pages={6734--6747},
  year={2024}
}

@article{yi2024jailbreak,
  title={Jailbreak attacks and defenses against large language models: A survey},
  author={Yi, Sibo and Liu, Yule and Sun, Zhen and Cong, Tianshuo and He, Xinlei and Song, Jiaxing and Xu, Ke and Li, Qi},
  journal={arXiv preprint arXiv:2407.04295},
  year={2024}
}

@article{ferrag2025prompt,
  title={From prompt injections to protocol exploits: Threats in LLM-powered AI agents workflows},
  author={Ferrag, Mohamed Amine and Tihanyi, Norbert and Hamouda, Djallel and Maglaras, Leandros and Lakas, Abderrahmane and Debbah, Merouane},
  journal={ICT Express},
  year={2025},
  publisher={Elsevier}
}

@misc{he2021debertav3,
  title={DeBERTaV3: Improving DeBERTa using ELECTRA-Style Pre-Training with Gradient-Disentangled Embedding Sharing},
  author={Pengcheng He and Jianfeng Gao and Weizhu Chen},
  year={2021},
  eprint={2111.09543},
  archivePrefix={arXiv},
  primaryClass={cs.CL}
}

@inproceedings{williams2018broad,
  title={A broad-coverage challenge corpus for sentence understanding through inference},
  author={Williams, Adina and Nangia, Nikita and Bowman, Samuel},
  booktitle={Proceedings of the 2018 Conference of the North American Chapter of the Association for Computational Linguistics: Human Language Technologies, Volume 1 (Long Papers)},
  pages={1112--1122},
  year={2018}
}

@article{wei2025memguard,
  title={A-memguard: A proactive defense framework for llm-based agent memory},
  author={Wei, Qianshan and Yang, Tengchao and Wang, Yaochen and Li, Xinfeng and Li, Lijun and Yin, Zhenfei and Zhan, Yi and Holz, Thorsten and Lin, Zhiqiang and Wang, XiaoFeng},
  journal={arXiv preprint arXiv:2510.02373},
  year={2025}
}

@article{sunil2026memory,
  title={Memory Poisoning Attack and Defense on Memory Based LLM-Agents},
  author={Sunil, Balachandra Devarangadi and Sinha, Isheeta and Maheshwari, Piyush and Todmal, Shantanu and Mallik, Shreyan and Mishra, Shuchi},
  journal={arXiv preprint arXiv:2601.05504},
  year={2026}
}

@article{bhardwaj2026superlocalmemory,
  title={SuperLocalMemory: Privacy-preserving multi-agent memory with Bayesian trust defense against memory poisoning},
  author={Bhardwaj, Varun Pratap},
  journal={arXiv preprint arXiv:2603.02240},
  year={2026}
}
